\documentclass[final]{anthology-ch} 

\usepackage{booktabs}
\usepackage{graphicx}
\usepackage[normalem]{ulem} 

\usepackage[most]{tcolorbox}
\usepackage{enumitem}
\usepackage{tikz}
\usepackage{fontawesome5}  
\tcbuselibrary{skins,hooks}
\usetikzlibrary{arrows.meta, positioning, decorations.markings}
\usepackage[table]{xcolor}
\usepackage{colortbl}
\usepackage{pgfplots}
\usepackage{caption} 
\usepackage{csquotes}
\MakeOuterQuote{"} 

\title{Estranged Predictions: Measuring Semantic Category Disruption with Masked Language Modelling}

\author[1]{Yuxuan Liu}[
  orcid=0009-0005-4960-1610
]

\author[2]{Haim Dubossarsky}[
  orcid=0000-0002-2818-6113
]

\author[1]{Ruth Ahnert}[
  orcid=0000-0002-8503-1580
]

\affiliation{1}{School of Arts, Queen Mary University of London, London, United Kingdom}
\affiliation{2}{School of Electronic Engineering and Computer Science, Queen Mary University of London, London, United Kingdom}


\begin{document}

\maketitle
\begin{abstract}
This paper examines how science fiction destabilises ontological categories by measuring conceptual permeability across the terms \textit{human}, \textit{animal}, and \textit{machine} using masked language modelling (MLM). Drawing on corpora of science fiction (Gollancz SF Masterworks) and general fiction (NovelTM), we operationalise Darko Suvin's theory of estrangement as computationally measurable deviation in token prediction, using RoBERTa to generate lexical substitutes for masked referents and classifying them via Gemini. We quantify conceptual slippage through three metrics: retention rate, replacement rate, and entropy, mapping the stability or disruption of category boundaries across genres. Our findings reveal that science fiction exhibits heightened conceptual permeability, particularly around \textit{machine} referents, which show significant cross-category substitution and dispersion. \textit{Human} terms, by contrast, maintain semantic coherence and often anchor substitutional hierarchies. These patterns suggest a genre-specific restructuring within anthropocentric logics. We argue that estrangement in science fiction operates as a controlled perturbation of semantic norms, detectable through probabilistic modelling, and that MLMs, when used critically, serve as interpretive instruments capable of surfacing genre-conditioned ontological assumptions. This study contributes to the methodological repertoire of computational literary studies and offers new insights into the linguistic infrastructure of science fiction.
\end{abstract}

\section{Introduction}
Science fiction has long served as a speculative mirror for our assumptions about identity and existence, and as a crucible in which such categories are actively contested, dismantled, and reconstituted. The boundary between \textit{human} and \textit{Other} emerges not as a stable demarcation but as a zone of negotiation, hybridisation, and semantic leakage. In this space, estrangement functions not only as a narrative device but as a cognitive operation, one that defamiliarises hegemonic epistemologies and opens possibilities for alternative recognition and disidentification \cite{Suvin1979}. 

This ontological disturbance finds its linguistic correlate in the destabilisation of language itself, where familiar referents are defamiliarised and perception is slowed into reflective, non-automatic engagement, a process theorised in Viktor Shklovsky's concept of defamiliarisation \cite{Shklovsky2017}. Estrangement thus operates not only at the level of macro-narrative architecture, but at the micro-level of syntax and lexical substitution. 

While substantial critical work has illuminated the speculative imagination of the \textit{posthuman} \cite{Braidotti2013}, the \textit{cyborg} \cite{Haraway1991}, and the \textit{animal} \cite{Wolfe2010}, these contributions have tended to privilege philosophical, ethical, or narratological perspectives. Less attention has been paid to the linguistic structure through which ontological categories are produced, troubled, or undone. A solely thematic approach risks obscuring the microstructural mechanisms, such as syntactic juxtapositions, semantic proximities, and predictive patterns of substitution, by which science fiction reconfigures the conceptual intelligibility of alterity, for such readings remain predominantly qualitative in scope and ill-equipped to chart statistical patterns across large corpora.

This study therefore poses three interrelated questions: How is the conceptual boundary surrounding the \textit{human} rendered porous or unstable in literary language? With which kinds of entities, including \textit{animal}, \textit{machine}, or others, is this permeability most frequently negotiated? And how do these patterns of semantic proximity vary between speculative and non-speculative fiction? These questions are critical for rethinking how science fiction not only narrates difference, but actively reorganises the linguistic scaffolding through which difference is rendered legible.

\section{Background and Related Work}
Distributional semantics offers a powerful framework for interrogating such phenomena. As J. R. Firth famously proposed, "you shall know a word by the company it keeps" \cite{Firth1957}, a foundational principle further developed by Zellig Harris \cite{Harris1954} and formalised in vector-based models of semantics \cite{Turney2010,Baroni2014}. From this perspective, semantic categories such as \textit{human}, \textit{animal}, and \textit{machine} are not defined by essential features but emerge through mutable contextual associations that vary by genre, historical moment, and discursive environment \cite{underwood2021mapping, klein2025provocations}. 

Recent advances in language modelling, particularly through contextualised models such as BERT \cite{Devlin2019} and RoBERTa \cite{Liu2019}, have expanded the empirical tools available for tracking these dynamics. Unlike traditional distributional methods, either static embeddings or those that rely on raw co-occurrence frequencies, contextualised models can predict masked or "missing" tokens based on their sentence context in a paradigm called masked language modelling (MLM). This experimental paradigm allows for the reconstruction of latent semantic expectations and substitution probabilities in sentence-level contexts. When applied to science fiction, MLMs reveal how referential expectations, for example, around \textit{human}, \textit{machine}, and \textit{animal} shift across discursive regimes, and how meaning becomes unstable in moments of predictive uncertainty.

The \textit{Living with Machines} (LwM) project has demonstrated the potential of MLMs to surface latent tensions in language use, particularly by repurposing a characteristic property of large-scale language models, namely their tendency to default to high-probability, statistically dominant completions, as an analytical lens through which to detect linguistic departures from normativity. In this context, the project explored how language models could be used to detect instances of linguistic usage that would appear "surprising" to a model trained on a specific historical corpus, particularly in relation to depictions of animate machines \cite{CollArdanuy2020,Wilson2023}.  

\section{Approach}
This study builds on the approach employed in the LwM project. Drawing from a corpus of nineteenth-century texts, the LwM team selected sentences concerning \textit{machines} and masked the \textit{machine}-related terms. They then prompted a historical language model that they had fine-tuned \cite{Hosseini2021}, to generate probable lexical substitutions for the masked words, as in their first example:

\begin{quote}
    \textbf{Original sentence}: And why should one say that the machine does not live?
    
    \textbf{Masked sentence}: And why should one say that the {\small\texttt{[MASK]}} does not live?
    
    \textbf{Predictions with scores}: \textit{man} (5.0788), \textit{person} (4.4484), \textit{other} (4.1866), \textit{child} (4.1600), \textit{king} (4.1510), \textit{patient} (4.1249), \textit{one} (4.1141), \textit{stranger} (4.1067), ... 
\end{quote} 

The method provided a powerful way of detecting atypical animacy, as well as revealing deep-seated cultural and ideological biases embedded within the nineteenth-century training corpus. \cite{Wilson2023}.

Building on this precedent, our study extends the application of MLM to a comparative corpus comprising science fiction and general fiction. It focuses on the conceptual permeability of three ontological categories, namely \textit{human}, \textit{animal}, and \textit{machine}, by masking these terms and examining the substitutions proposed by a contemporary language model. By analysing what a probabilistic model deems plausible within given linguistic contexts, this study captures not only the stability of referential expectations but also their points of breakdown. These misalignments gesture toward areas of semantic permeability or categorical ambiguity. Taken collectively, these "errors" reveal broader patterns of generic difference both within science fiction, and between the genre and the broader fiction landscape, thereby exposing genre-specific estrangement effects that operate beneath the level of explicit narration.

Through this comparative framework, our study operationalises the notion of estrangement through measurable shifts in prediction probabilities generated by a masked language model. Specifically, while it may seem self-evident that science fiction, by virtue of its genre-specific discourse, engenders increased conceptual permeability across three core ontological categories of \textit{human}, \textit{animal}, and \textit{machine}, we ask whether this assumption can be substantiated at scale, and, if so, what our analytical pipeline can reveal about the mechanisms by which these ontological boundaries are explored at the microstructural level of language. 

\section{Materials and methods}

\subsection{Corpus Selection}
To enable a controlled comparison of conceptual permeability across science fiction and general fiction, two corpora were selected for their viability as bounded discursive environments within which the linguistic negotiation of categorical boundaries could be meaningfully modelled. The science fiction corpus comprises 336 published works drawn from the Gollancz SF Masterworks series (1818–2019), encompassing both standalone novels and individually extracted stories from anthologies, with over 90\% of the material concentrated between 1910 and 2000. The general fiction corpus is derived from the HathiTrust NovelTM dataset \cite{Underwood2020}, from which we randomly sampled 700 Anglophone works published between 1910 and 2000.

\subsection{Contextualised Model}
We employed \textit{roberta-base} \cite{Liu2019} to generate predictions for masked tokens across all sentences. RoBERTa (A Robustly Optimised BERT Pretraining Approach) is a refined implementation of BERT \cite{Devlin2019}, a bidirectional transformer designed to learn the probabilistic relationships between words by predicting masked tokens in context; RoBERTa is conceptualised in this study as an instrument for detecting estrangement through which to detect, replicate, and reflect the distributional conventionalities of linguistic production as sedimented through pragmatic histories and examine the structural inertia of language: its anthropocentric defaults, its resistance to nonhuman agency, and its syntactic regulation of who or what may occupy grammatically legitimate positions of action.

\subsection{Experimental Procedure}
\begin{figure*}[htbp]
\centering
\scalebox{0.7}{
\begin{tikzpicture}[
    corpusbox/.style={rectangle, rounded corners=5pt, minimum width=5.5cm, minimum height=1cm, 
                      draw=black!40, fill=green!12, line width=1.2pt, font=\large\bfseries, 
                      align=center},
    processbox/.style={rectangle, rounded corners=5pt, minimum width=11cm, minimum height=1cm,
                       draw=black!60, line width=1.5pt, font=\large\bfseries, align=center, 
                       dashed, dash pattern=on 4pt off 3pt, fill=white},
    categorybox/.style={rectangle, rounded corners=4pt, minimum width=2.8cm, minimum height=0.9cm,
                        line width=1.3pt, font=\large\bfseries, align=center},
    predbox/.style={rectangle, rounded corners=4pt, minimum width=1.8cm, minimum height=0.7cm,
                    line width=1pt, font=\normalsize\bfseries, align=center},
    arrow/.style={-{Stealth[length=10pt, width=7pt]}, line width=2.2pt, draw=black!70},
    thickarrow/.style={-{Stealth[length=12pt, width=8pt]}, line width=3pt, draw=black!70}
]

\definecolor{machinecolor}{RGB}{255, 200, 100}
\definecolor{humancolor}{RGB}{180, 235, 200}
\definecolor{animalcolor}{RGB}{255, 180, 180}

\node[corpusbox] (scifi) at (-3.5, 0) {Science Fiction Corpus};
\node[corpusbox] (genfic) at (3.5, 0) {General Fiction Corpus};

\node[processbox] (extract) at (0, -2) {Sentence Extraction};

\draw[arrow] (scifi.south) -- (extract.north);
\draw[arrow] (genfic.south) -- (extract.north);

\node[categorybox, fill=machinecolor, draw=orange!70] (machine) at (-3.5, -4) {
    \textcolor{orange!70}{\faRobot}\ Machine
};
\node[categorybox, fill=humancolor, draw=teal!70] (human) at (0, -4) {
    \textcolor{teal!70}{\faUser}\ Human
};
\node[categorybox, fill=animalcolor, draw=red!60] (animal) at (3.5, -4) {
    \textcolor{red!60}{\faPaw}\ Animal
};

\draw[arrow] (extract.south) -- (machine.north);
\draw[arrow] (extract.south) -- (human.north);
\draw[arrow] (extract.south) -- (animal.north);

\node[processbox] (masking) at (0, -6) {Entity Masking};

\draw[arrow] (machine.south) -- (masking.north);
\draw[arrow] (human.south) -- (masking.north);
\draw[arrow] (animal.south) -- (masking.north);

\node[processbox, text width=11cm] (roberta) at (0, -8.2) {
    \textbf{RoBERTa-base Masked Language Modeling}\\
    \textbf{(Top-5 predictions)}
};

\draw[thickarrow] (masking.south) -- (roberta.north);

\node[predbox, fill=red!30, draw=red!60] (pred1) at (-5, -10.2) {Pred1};
\node[predbox, fill=orange!30, draw=orange!60] (pred2) at (-2.5, -10.2) {Pred2};
\node[predbox, fill=green!30, draw=green!60] (pred3) at (0, -10.2) {Pred3};
\node[predbox, fill=cyan!30, draw=cyan!60] (pred4) at (2.5, -10.2) {Pred4};
\node[predbox, fill=blue!30, draw=blue!60] (pred5) at (5, -10.2) {Pred5};

\foreach \x in {pred1, pred2, pred3, pred4, pred5} {
    \draw[arrow] (roberta.south) -- (\x.north);
}

\node[processbox] (gemini) at (0, -12.2) {Gemini API Semantic Classification};

\foreach \x in {pred1, pred2, pred3, pred4, pred5} {
    \draw[arrow] (\x.south) -- (gemini.north);
}

\end{tikzpicture}
}
\caption{Cross-genre conceptual permeability detection model}
\label{fig:conceptual_permeability_model}
\end{figure*}
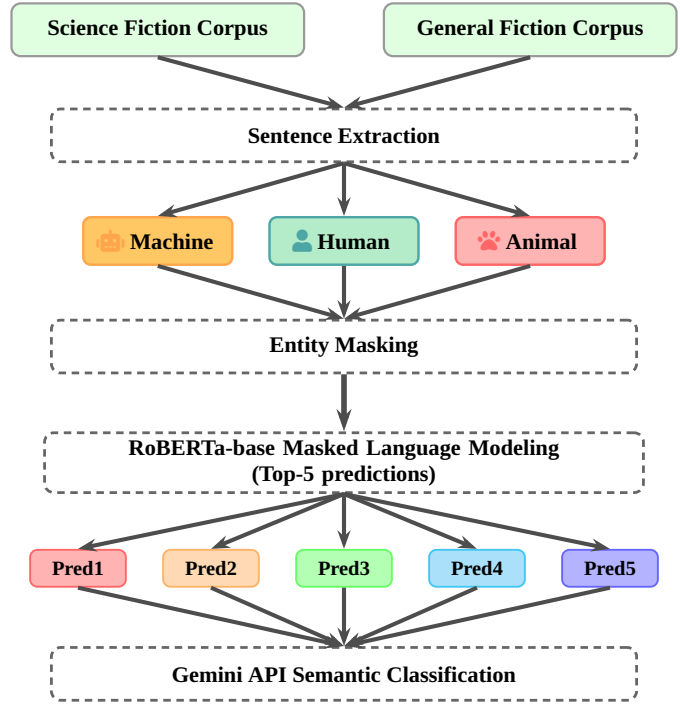

Our experimental procedure consists of three core stages, illustrated in Figure~\ref{fig:conceptual_permeability_model}.\footnote{Development used Jupyter Notebooks launched via the OnDemand environment \cite{Hudak2018}, with classification via the Gemini API. All code for the pipeline is available at \url{https://github.com/yuxliuu89/semantic-category-disruption-mlm}.}

\subsubsection{Sentence Extraction and Masked Language Modelling}
A case-insensitive lexical query extracts all sentences containing the target terms \textit{human/humans}, \textit{human being/human beings}, \textit{animal/animals}, and \textit{machine/machines} from both corpora. This procedure yields 11,709 sentences from the science fiction corpus and 8972 sentences from general fiction corpus. In each sentence, the target lexical item is replaced with a {\small\texttt{[MASK]}} token to prepare the input for MLM. Some of these sentences contain more than one target category; for instance, a sentence might mention both a \textit{human} and a \textit{machine}, or even include all three. In such cases, the sentence is processed iteratively: in each pass, only one term is masked, while all others remain intact. Thus, if a sentence contains two target terms, the model is run twice on the same sentence: once with the first term masked and the second visible, and once with the reverse configuration. As a result, the total number of masked sentences exceeds the number of original sentences. During prediction extraction, the model returns its top 5 candidate predictions, for each sentence, along with associated token-level probabilities. 
 
\subsubsection{Semantic Classification via Gemini}
The model's predictions are subjected to semantic classification using Google's Gemini \cite{GoogleGemini2.5_2025}. Each predicted token is classified within its full sentential context via a prompt comprising the masked sentence and the model's prediction. The classification framework employs a taxonomy centred on three core ontological categories: \textit{Human}, \textit{Animal}, and \textit{Machine}. Each core category is supplemented by adjacent figures of \textit{otherness}, including \textit{Other-Human}, \textit{Other-Animal}, and \textit{Other-Machine}, as well as further categories such as \textit{Other-Hybrid} and \textit{Other-Ambiguous}, the latter of which accounts for liminal or indeterminate referents. Additional categories are generated ad hoc by the Gemini model in response to the semantic type of each MLM prediction and its contextual specificity. All predictions are processed in batches of 200, with quality control mechanisms in place to ensure a classification coverage rate exceeding 98\%. To mitigate category fragmentation and overlap, Gemini's post-classification fusion is employed to consolidate semantically adjacent labels both within and across ontological domains.

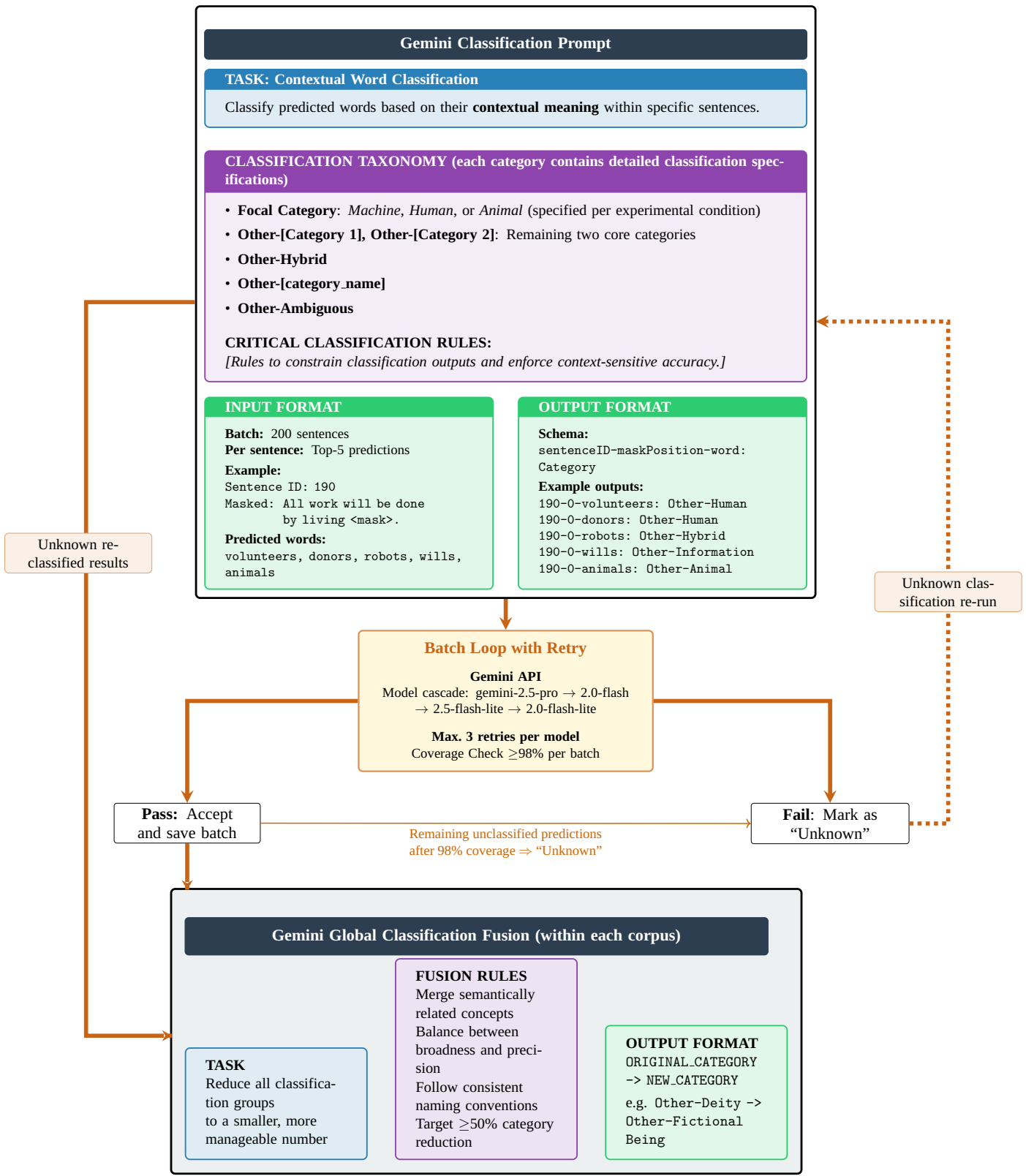
\begin{figure*}[htbp]
\makebox[\textwidth][l]{\hspace*{-1.5cm}\scalebox{0.7}{
\begin{tikzpicture}[
    mainbox/.style={draw=black, line width=1.5pt, rounded corners=3pt, fill=white, text width=15.5cm, inner sep=6pt, align=left},
    qcnode/.style={draw=black, fill=white, rounded corners=2pt, minimum height=0.6cm, align=center, font=\footnotesize},
    fusionbox/.style={draw=black, line width=1.5pt, rounded corners=3pt, fill=fusionbg, text width=15.5cm, minimum height=5.5cm, inner sep=10pt},
    smallnote/.style={draw=qcdark!60, fill=qcdark!10, rounded corners=3pt, font=\footnotesize, align=center, inner sep=4pt},
    arrow/.style={-stealth, line width=3pt, draw=qcdark}
]

\definecolor{taskcolor}{RGB}{41,128,185}
\definecolor{taxonomycolor}{RGB}{142,68,173}
\definecolor{iocolor}{RGB}{46,204,113}
\definecolor{qcdark}{RGB}{200,100,20}
\definecolor{fusionbg}{RGB}{236,240,241}
\definecolor{titlecolor}{RGB}{44,62,80}
\definecolor{qcbg}{RGB}{255,248,220}

\node[mainbox, xshift=-1cm] (main) {
    \begin{tcolorbox}[enhanced, colback=titlecolor, colframe=titlecolor, width=15.5cm, arc=1mm, boxrule=0pt,
        fontupper=\color{white}\large\bfseries, halign=center, top=1mm, bottom=1mm]
        Gemini Classification Prompt
    \end{tcolorbox}

    \begin{tcolorbox}[enhanced, colback=taskcolor!15, colframe=taskcolor, width=\textwidth,
        arc=1mm, boxrule=1pt, title={\textbf{TASK: Contextual Word Classification}},
        fonttitle=\bfseries\normalsize, coltitle=white, colbacktitle=taskcolor]
        \normalsize Classify predicted words based on their \textbf{contextual meaning} within specific sentences.
    \end{tcolorbox}

    \vspace{1mm}

    \begin{tcolorbox}[enhanced, colback=taxonomycolor!10, colframe=taxonomycolor, width=\textwidth,
        arc=1mm, boxrule=1pt,
        title={\textbf{CLASSIFICATION TAXONOMY (each category contains detailed classification specifications)}},
        fonttitle=\bfseries\normalsize, coltitle=white, colbacktitle=taxonomycolor]
        \normalsize
        \begin{itemize}[leftmargin=*, itemsep=0pt]
            \item \textbf{Focal Category}: \textit{Machine}, \textit{Human}, or \textit{Animal} (specified per experimental condition)
            \item \textbf{Other-[Category 1], Other-[Category 2]}: Remaining two core categories
            \item \textbf{Other-Hybrid}
            \item \textbf{Other-[category\_name]}
            \item \textbf{Other-Ambiguous}
        \end{itemize}
        \textbf{CRITICAL CLASSIFICATION RULES:}\\ \textit{[Rules to constrain classification outputs and enforce context-sensitive accuracy.]}
    \end{tcolorbox}

    \vspace{1mm}

    \begin{minipage}[t]{0.48\textwidth}
        \begin{tcolorbox}[enhanced, colback=iocolor!15, colframe=iocolor, arc=1mm, boxrule=1pt,
            height=5cm, title={\textbf{INPUT FORMAT}}, fonttitle=\bfseries\normalsize, coltitle=white, colbacktitle=iocolor]
            \small
            \textbf{Batch:} 200 sentences\\
            \textbf{Per sentence:} Top-5 predictions\\[1mm]
            \textbf{Example:}\\
            \texttt{Sentence ID: 190}\\
            \texttt{Masked: All work will be done}\\
            \texttt{\phantom{Masked: }by living <mask>.}\\[1mm]
            \textbf{Predicted words:}\\
            \texttt{volunteers, donors, robots, wills, animals}
        \end{tcolorbox}
    \end{minipage}
    \hfill
    \begin{minipage}[t]{0.48\textwidth}
        \begin{tcolorbox}[enhanced, colback=iocolor!15, colframe=iocolor, arc=1mm, boxrule=1pt,
            height=5cm, title={\textbf{OUTPUT FORMAT}}, fonttitle=\bfseries\normalsize, coltitle=white, colbacktitle=iocolor]
            \small
            \textbf{Schema:}\\
            \texttt{sentenceID-maskPosition-word: Category}\\[1mm]
            \textbf{Example outputs:}\\
            \texttt{190-0-volunteers: Other-Human}\\
            \texttt{190-0-donors: Other-Human}\\
            \texttt{190-0-robots: Other-Hybrid}\\
            \texttt{190-0-wills: Other-Information}\\
            \texttt{190-0-animals: Other-Animal}
        \end{tcolorbox}
    \end{minipage}
};

\node[below=0.8cm of main, draw=qcdark!70, line width=1.2pt, rounded corners=3pt, 
      fill=qcbg, text width=7cm, inner sep=8pt, align=center, font=\small] (qc) {
{\large\textbf{\color{qcdark}Batch Loop with Retry}}\\[3mm]
\textbf{Gemini API}\\
Model cascade: gemini-2.5-pro $\rightarrow$ 2.0-flash\\
$\rightarrow$ 2.5-flash-lite $\rightarrow$ 2.0-flash-lite\\[3mm]
\textbf{Max.\ 3 retries per model}\\
Coverage Check $\geq$98\% per batch
};

\node[qcnode, below left=0.8cm and 2.5cm of qc, text width=3.5cm, font=\large] (pass) {%
\textbf{Pass:} Accept and save batch};

\node[qcnode, below right=0.8cm and 2.5cm of qc, text width=3.8cm, font=\large] (fail) {%
\textbf{Fail}: Mark as "Unknown"};

\draw[->, thick, color=orange!80!black]
  (pass.east) --
  node[midway, below, font=\small, text=orange!80!black, align=center]
  {Remaining unclassified predictions\\after 98\% coverage $\Rightarrow$ "Unknown"}
  (fail.west);

\node[fusionbox, below=3cm of qc, xshift=-0.5cm] (fusion) {
    \begin{tcolorbox}[enhanced, colback=titlecolor, colframe=titlecolor, width=15cm, arc=1mm,
        boxrule=0pt, fontupper=\color{white}\large\bfseries, halign=center]
        Gemini Global Classification Fusion (within each corpus)
    \end{tcolorbox}

    \vspace{-2mm}
    \begin{minipage}[t]{4.7cm}
        \begin{tcolorbox}[enhanced, colback=taskcolor!15, colframe=taskcolor, arc=1mm, boxrule=1pt]
            \normalsize\textbf{TASK}\\
            Reduce all {classification groups}\\
            to a smaller, more manageable number
        \end{tcolorbox}
    \end{minipage}
    \hfill
    \begin{minipage}[t]{4.7cm}
        \begin{tcolorbox}[enhanced, colback=taxonomycolor!15, colframe=taxonomycolor, arc=1mm, boxrule=1pt]
            \normalsize\textbf{FUSION RULES}\\
            Merge semantically related concepts\\
            Balance between broadness and precision\\
            Follow consistent naming conventions\\
            Target $\geq$50\% category reduction
        \end{tcolorbox}
    \end{minipage}
    \hfill
    \begin{minipage}[t]{4.7cm}
        \begin{tcolorbox}[enhanced, colback=iocolor!15, colframe=iocolor, arc=1mm, boxrule=1pt]
            \normalsize\textbf{OUTPUT FORMAT}\\
            \texttt{ORIGINAL\_CATEGORY -> NEW\_CATEGORY}\\[1mm]
            e.g. \texttt{Other-Deity -> Other-Fictional Being}
        \end{tcolorbox}
    \end{minipage}
};

\draw[arrow] (main) -- (qc);
\draw[arrow] (qc) -| (pass);
\draw[arrow] (qc) -| (fail);

\draw[line width=3pt, draw=qcdark] 
  (main.west) -- ++(-2.8cm,0) -- ++(0,-19cm) -- ++(2cm,0)
  node[at end, xshift=3pt] {\tikz \draw[-{Stealth[length=7pt,width=8pt]}, qcdark, line width=3pt] (0,0)--(0.001,0);};

\draw[line width=3pt, draw=qcdark] 
  (pass.south) -- ++(0,-1cm)
  node[at end, yshift=-3pt] {\tikz \draw[-{Stealth[length=7pt,width=8pt]}, qcdark, line width=3pt] (0,0)--(0,-0.001);};

\draw[arrow, dashed] (fail.east) -- ++(1cm,0) -- ++(0,13cm) -- ++(-3.4cm,0);

\node[smallnote, font=\normalsize, align=center, text width=3.5cm]
  at ([xshift=1cm, yshift=6cm]fail.east)
  {Unknown classification re-run};

\node[smallnote, font=\normalsize, align=center, text width=3.5cm]
  at ([xshift=-3cm, yshift=-6.5cm]main.west)
  {Unknown reclassified results};

\end{tikzpicture}
}}

\caption{Three-stage Gemini-based classification workflow}
\label{fig:gemini_classification_workflow}
\end{figure*}

\subsection{Metrics of Analysis}

The interpretive framework is structured around three related metrics: 

\subsubsection{Retention Rate}
This metric assesses the extent to which the model's predicted substitutes for a masked word remain within the same semantic category as the original term. For example, if the masked word \textit{human} is replaced by terms such as \textit{person}, \textit{man}, or \textit{child}, the substitution is considered category-preserving. Retention rate thus serves as an index of category stability, reflecting how consistently the model maintains the semantic identity of key terms during prediction.

\subsubsection{Replacement Rate}
This metric quantifies how often the model's predictions transgress the semantic boundaries delineating the categories of \textit{human}, \textit{animal}, and \textit{machine}. Specifically, the replacement rate indexes instances in which a masked lexical item originally classified within one category is predicted to belong to another, such as when a term denoting a \textit{machine} elicits top-ranked predictions associated with \textit{human}, \textit{animal}, or \textit{deity}. These cross-category substitutions are treated as markers of conceptual permeability. A higher replacement rate thus signals greater semantic fluidity and ontological instability, whereas lower rates suggest the reinforcement of categorical boundaries. Moreover, by attending to the directional asymmetries in replacement patterns (e.g., \textit{machine} $\rightarrow$ \textit{human} versus \textit{human} $\rightarrow$ \textit{machine}), this metric allows for an analysis of how specific genres modulate the movement of meaning across categories of being, and whether, and in which direction, they foster slippage, containment, or reification in the linguistic construction of subjecthood.

\subsubsection{Entropy}
Entropy is used here to quantify the degree of uncertainty in the model's predictions. As a measure of probability dispersion, entropy reflects how the model navigates semantic constraints embedded within different literary contexts. Low entropy values correspond to highly concentrated distributions, where the model assigns disproportionately high probability to a single lexical candidate, suggesting strong contextual anchoring and reduced semantic ambiguity. Conversely, high entropy signals a more evenly distributed probability mass across multiple candidates, indicating a looser context that invites several plausible predictions. While the previous metrics evaluated the model's predictions by assessing whether each masked token was retained within or replaced across its original category, entropy was used to capture the probabilistic dispersion across the top five predictions collectively, thereby reflecting how uncertainty is distributed within the model's semantic space.

\subsubsection{Aggregation Schemes} 
In quantifying the retention of a masked token within its original semantic category and its replacement into alternative categories, two ways of aggregating RoBERTa's top-5 predictions for each masked position were considered. Method~1 treats the five highest-probability predictions for each masked position as equally weighted candidates, interpreting their presence as an indication of semantic plausibility. For example, a prediction for \textit{machine} with a probability of 0.02 is counted in the same way as one with a probability of 0.35. Under this method, retention is defined as:
\[
\text{Retention} = \frac{\text{Number of source-category predictions}}{\text{Total number of predictions}},
\]
with analogous calculations applied to each replacement category.

By contrast, Method~2 weights candidates according to their probabilities, summing the total probability mass assigned to each category within a sentence and then averaging across all masked positions. Retention is thus calculated as:
\[
\text{Retention} = \frac{1}{n}\sum_{\substack{\text{source} \\ \text{category}}} P_{(x)},
\]
where \( P_{(x)} \) denotes the probability assigned by the model to a token belonging to the source category and \textit{n} represents the number of masked positions.

However, Method~1 captures weak but non-trivial cross-category substitutions, amplifying marginal predictions that may index latent semantic permeability. By granting uniform status to all top-5 predictions irrespective of their associated probabilities, Method~1 lowers the threshold for registering both category-consistent and category-divergent outcomes, ensuring that incursions across semantic boundaries are not dismissed prematurely. Given that the analytical aim is to detect the semantic permeability of ontological categories rather than to replicate the model's internal probability structure, and in view of the high consistency observed between the retention and replacement patterns generated by Method~1 and Method~2 (full results for Method~2 are provided in the appendix; see Figure~\ref{fig:combined_top10_method2}), the use of an equal-weight approach that retains marginal signals under these conditions remains analytically justified.

\subsubsection{Statistical Tests} 
For retention, we binarised the model's top-5 predictions, whether they remain within the same semantic category or not, for each masked sentence, and expressed this as a proportion (0–1). To test whether retention differs between corpora, we computed the observed mean difference in the binarised per-sentence retention between Gollancz SF and NovelTM, and evaluated its significance using a 10,000-iteration permutation test that randomly shuffled corpus labels. We also obtained 95\% confidence intervals for the mean difference by bootstrap resampling sentences within each corpus. This model-free approach ensures that significance does not depend on distributional assumptions and treats each sentence as an independent observation. For Entropy analysis, we used a two-way analysis of variance (ANOVA) test with category and corpus as the main variables and their interaction.

\section{Results}
\subsection{Conceptual Retention and Semantic Dispersion Across Genres}

\begin{figure}[htbp]
    \centering
    \includegraphics[width=\textwidth]{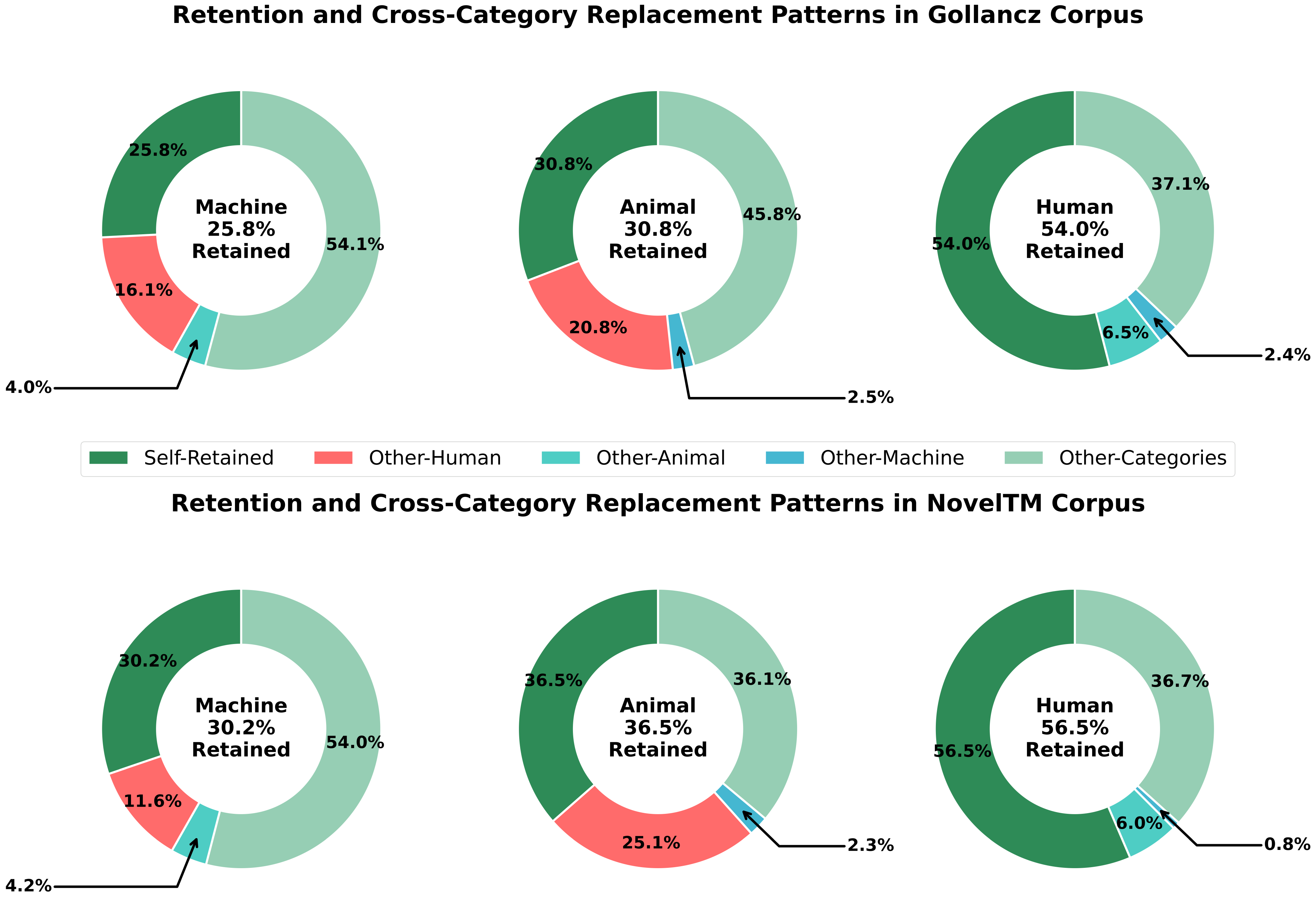}
    \caption{Cross-Entity Replacement Patterns in Gollancz SF and NovelTM Corpora}
    \label{fig:combined_donut}
\end{figure}

The computational findings presented in this subsection delineate a spectrum of conceptual stability and semantic dispersion across the categories \textit{human}, \textit{animal}, and \textit{machine}, as measured respectively by retention rates and entropy values in Gollancz SF and NovelTM (Figure~\ref{fig:combined_donut}). 

Among the three conceptual categories, \textit{machine} terms demonstrate the lowest rate of conceptual self-retention, indicating a marked semantic volatility. In the science fiction corpus, only 25.8\% of masked \textit{machine} tokens are predicted by RoBERTa as belonging to the same conceptual category, compared to 30.2\% in the general fiction corpus. This decline was statistically significant ($\Delta = -0.044$, 95\% CI [-0.063, -0.035], p < 0.001) in our permutation test, suggesting that science fiction narratives induce a destabilisation of the conceptual coherence of the \textit{machine} category.
The \textit{animal} category showed a similar pattern. In general fiction, 36.5\% of masked \textit{animal} tokens are predicted within the same category, a figure that drops to 30.8\% in science fiction ($\Delta = -0.056$, 95\% CI [-0.076, -0.045], p < 0.001).  

In contrast to the marked instability in the two categories above, the \textit{human} category exhibits a notable semantic resilience across genres. Retention rates remain comparatively high, with 54.0\% of masked \textit{human} tokens in science fiction predicted as belonging to the same conceptual category, compared to 56.5\% in general fiction. This modest decline did not reach statistical significance ($\Delta = -0.021$, 95\% CI [-0.045, -0.004], p = 0.1). This relative consistency suggests that the \textit{human} category maintains a high degree of semantic stability across both corpora, exhibiting limited permeability of conceptual boundaries when compared to other categories.

\begin{figure}[htbp]
    \centering 
    \makebox[\textwidth]{%
        \includegraphics[width=0.9\textwidth]{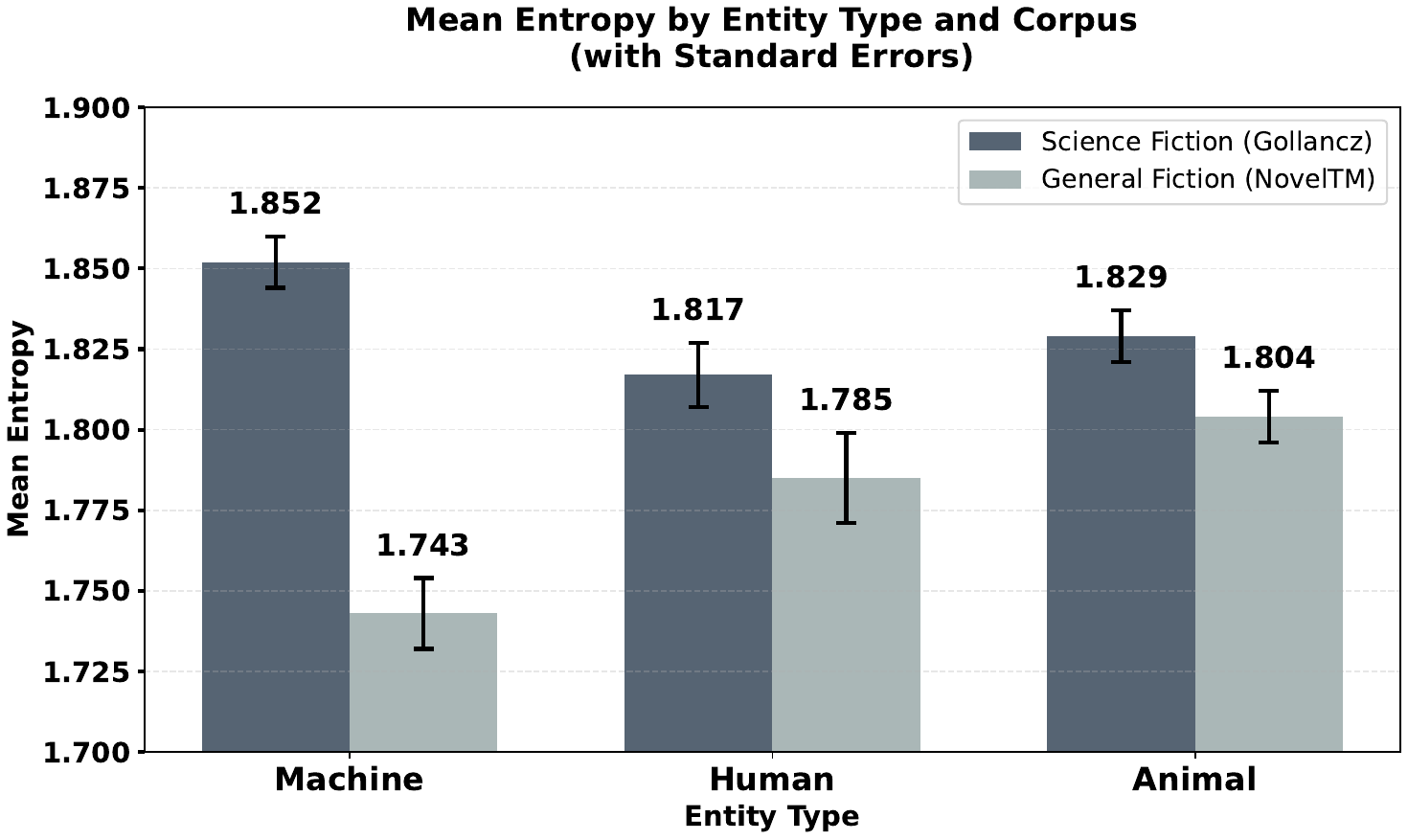}%
    }
    \caption{Mean Entropy by Entity Type and Corpus (with Standard Errors)}
    \label{fig:entropy_comparison}
\end{figure}

The entropy analysis substantiates and amplifies the findings drawn from retention rates by measuring the degree of predictive uncertainty associated with each conceptual category, namely the dispersion of plausible substitutes generated when the model is deprived of the original lexical item. Entropy analysis (Figure~\ref{fig:entropy_comparison}) shows a statistically significant corpus-by-category interaction ($F_{(2, 20652)}=15.7, p<.001$, $\eta$²=0.0015), indicating that overall entropy differs between corpora, but that the pattern of entropy variation across categories (\textit{human}, \textit{animal}, \textit{machine}) changes as a function of corpus type. In particular, masked \textit{machine} tokens in science fiction exhibiting the highest mean entropy, exceeding that of both \textit{animal} and \textit{human} referents. \footnote{The low $\eta$² reflects the inherent variability of sentence-level data rather than a weak effect. In linguistic datasets, small $\eta$² values are expected and still capture reliable, generalisable differences across conditions.} 

This elevated entropy in the science fiction corpus indicates that the \textit{machine} category elicits the widest semantic field, thereby marking it as the most conceptually unstable within that genre. The lower entropy observed in general fiction corpus, however, suggests that \textit{machine} remains more semantically constrained outside the science-fictional context. When compared across the two corpora, only \textit{machine}-related tokens register statistically significant entropy increase, from $H=1.743$ in NovelTM to $H=1.852$ in Gollancz SF ($t=8.93, p < .001$, Bonferroni post-hoc test). By contrast, the changes in entropy of either \textit{human} or \textit{animal} concepts did not reach statistical significance ($t=1.87$ and $t=2.04$, respectively). While this pattern broadly accords with existing expectations about science fiction's treatment of technology, it offers a statistical means of showing how that semantic pressure is enacted on machine referents in language, expanding their substitutional latitude and, by extension, destabilising their categorical fixity.

\subsection{Semantic Replacement and Conceptual permeability Across Genres}

\begin{figure}[htbp]
    \centering
    \makebox[\textwidth]{\includegraphics[width=1.2\textwidth]{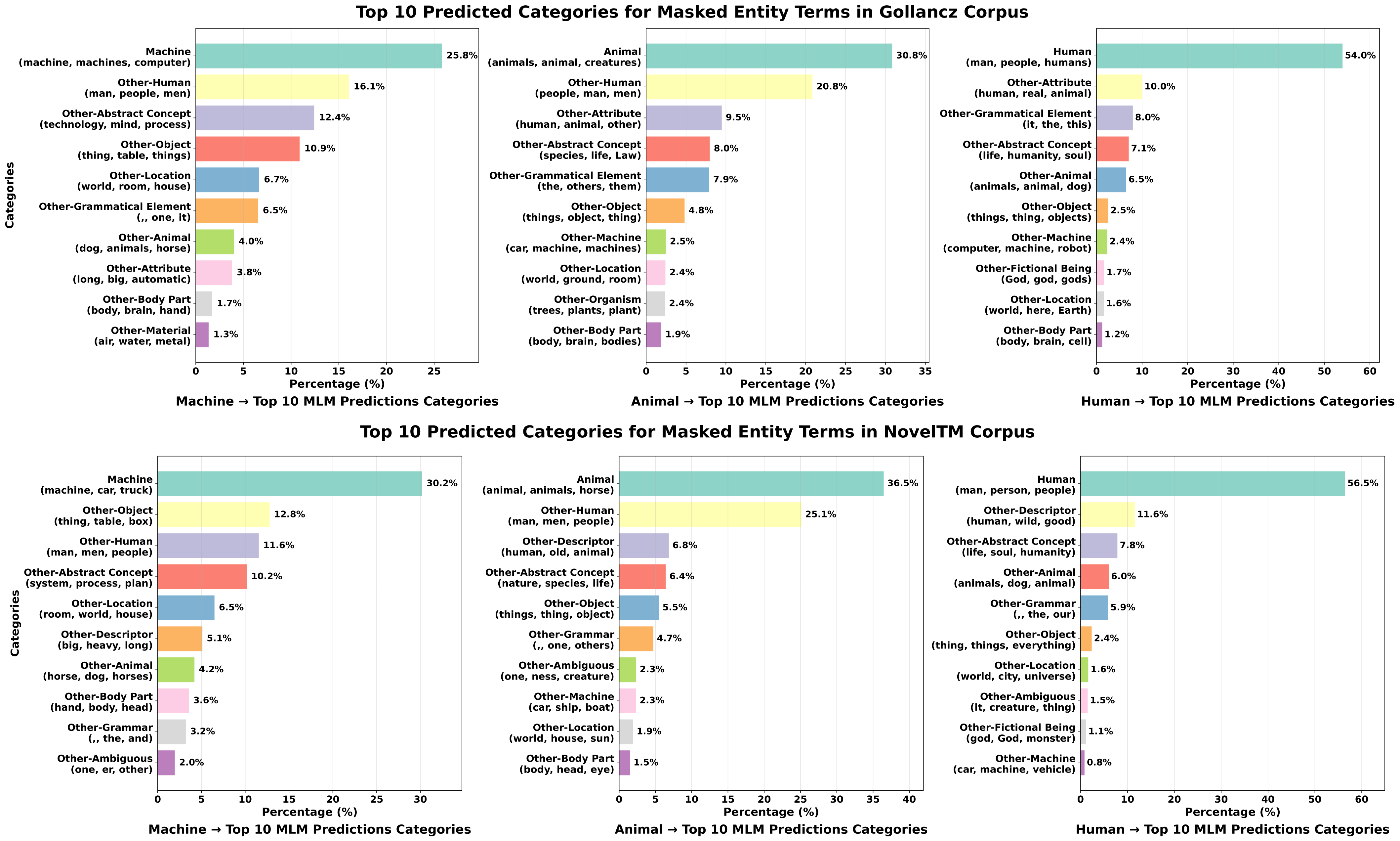}}
    \caption{Top 10 Predicted Categories for Masked Entity Terms in Gollancz SF and NovelTM Corpora}
    \label{fig:combined_top10_method1}
\end{figure}

Building upon the findings of the preceding subsection, this subsection examines how conceptual destabilisation manifests through replacement rate analysis (Figure~\ref{fig:combined_top10_method1}).

Comparison of replacement rates reveals a pronounced asymmetry in directional substitution across genres. In Gollancz SF, 16.1\% of masked \textit{machine} tokens were replaced with \textit{human}-related terms, while only 2.4\% of \textit{human} tokens were substituted with \textit{machine}-related ones. This directional imbalance is echoed in the NovelTM sample corpus, albeit at lower magnitudes, 11.6\% and 0.8\%, respectively, indicating that the anthropomorphisation of \textit{machines} is a widespread feature of fiction more generally, yet notably accentuated in Gollancz SF, where \textit{machine}-to-\textit{human} substitutions increase by approximately 38.8\%. 

The permeability of boundaries between \textit{human} and \textit{animal} categories also exhibits a marked asymmetry. In Gollancz SF, 6.5\% of \textit{human} tokens were replaced with \textit{animal}-related terms, while 20.8\% of \textit{animal} tokens were replaced by \textit{human} terms. NovelTM yields similar directional asymmetry: 6.0\% of \textit{human} tokens were replaced with \textit{animal} references, while 25.1\% of \textit{animal} tokens were supplanted by \textit{human} terms. This imbalance suggests that, in the NovelTM corpus, sentences containing \textit{animal} tokens more frequently align with the linguistic context in which RoBERTa expects \textit{human} referents to occur, revealing a more frequent tendency toward anthropomorphism. The Gollancz SF corpus, by contrast, exhibits this pattern less often. In the opposite direction, \textit{human} tokens in the Gollancz corpus are more often situated in contexts that resemble those in which RoBERTa anticipates \textit{animal} referents, indicating a more frequent inclination toward zoomorphism. Such an inversion in narrative perspective indicates a partial decentring of the human subject in Gollancz SF, a gesture towards post-anthropocentric thinking that, while not overriding the broader human-oriented defaults of language, nonetheless differentiates the genre in its narrative strategies.

The boundary between \textit{animal} and \textit{machine} categories also exhibits signs of permeability, albeit less markedly than in other category pairings. Despite the relatively low overall rates, the bidirectional substitutions observed in both corpora indicate a nuanced semantic overlap between zoological and mechanical domains. In Gollancz SF corpus, 2.5\% of masked \textit{animal} tokens were replaced with \textit{machine}-related terms, while 4.0\% of \textit{machine} tokens were substituted with \textit{animal}-related terms. The NovelTM corpus displays a comparable pattern, with a slightly lower replacement rate in the animal-to-machine direction (2.3\%) and a slightly higher rate in the machine-to-animal direction (4.2\%) compared to the Gollancz SF corpus. The directionality of these replacements reveals contrasting narrative tendencies across genres. Gollancz SF more frequently situates animals within machinic narratives, attributing mechanical qualities to animals; NovelTM, by contrast, more frequently describes machines through animal-related language, imbuing them with animacy.

\section{Discussion}
Across measures of retention, replacement, and entropy, Gollancz SF demonstrates a heightened degree of conceptual permeability and a higher incidence of semantic category disruption than NovelTM. These patterns indicate that the syntactic and semantic contexts surrounding the categories \textit{human}, \textit{animal}, and \textit{machine} are less constrained by the entrenched linguistic norms RoBERTa has internalised from standard English usage than those found in NovelTM. Silverstein has argued that these linguistic patterns are structured by a referential hierarchy which maps to degrees of perceived animacy  \cite{Silverstein1976}. In this hierarchy, the sequence proceeds down the scale from first- and second-person pronouns at the top, to third-person pronouns, then to proper names, human common nouns, other animates, and finally to inanimates \cite{Silverstein1976}. Our results reflect a partial adherence to this hierarchy: in both corpora, retention rates follow the order \textit{human} > \textit{animal} > \textit{machine}, consistent with the predicted ranking of animacy and agency. However, the patterns of cross-category replacement reveal reorganisitions of this ordering, with the degree of deviation differing across the two corpora. 

This reorganisation is more complex and fine-grained than the statistical overviews in our results section can convey. In the following discussion we show how RoBERTa's "wrong" predictions offer the key to more fine-grained insights. These can be navigated by adopting what Bamman et al. have termed \textit{classification-assisted close reading} \cite{Bamman2024} --- an approach that mobilises machine classification not as an end in itself, but as a supporting method of textual triage, directing critical attention to sites of conceptual instability. By classifying the more than one hundred thousand outputs generated by RoBERTa's MLM through Gemini, the analysis enables a targeted investigation of predictions. Such an approach enables us to trace a persistent pattern of conceptual permeability across the categories \textit{human}, \textit{animal}, and \textit{machine}: a phenomenon that resonates with theoretical accounts of posthuman technoculture, in which signification is no longer anchored to a stable referent but is instead distributed across a networked semantic field \cite{Hayles1999}.  

When \textit{machine} terms are replaced by human referents, the lexical contents of these substitutions reveal distinct genre-specific tendencies. In Gollancz SF, the five most frequent replacements for masked \textit{machine} tokens are \textit{man}, \textit{people}, \textit{woman}, \textit{human}, and \textit{guy}, each encompassing both singular and plural forms. These predictions indicate a gendered and age-coded reconfiguration of machinic entities. A notable feature is RoBERTa's frequent prediction of \textit{human} itself, rather than gender-signifying nouns such as \textit{man} or \textit{woman}. This pattern suggests that, compared with NovelTM, Gollancz SF more often assimilates \textit{machines} to the superordinate \textit{human} category without specifying a finer-grained social role. In contrast, while the NovelTM corpus also registers human-subcategorical replacements such as \textit{man} and \textit{woman}, its broader distribution gravitates toward functionally defined or service-oriented role terms including \textit{soldier}, \textit{driver}, and \textit{pilot}. The divergence between the two corpora thus signals distinct conceptions of linguistic agency: Gollancz SF reimagines the \textit{machine} as a bearer of subjectivity grounded in \textit{human} social and affective registers, whereas NovelTM constructs it as an extension of \textit{human} labour and function.

That subjectivity, however, is not monolithic. The model reveals that the humanisation of the \textit{machine} across the Gollancz SF corpus also disrupts stereotypical interactional norms, which are surfaced through the different sub-categories of \textit{human} entities predicted by RoBERTa for \textit{machine}. For example, in John O'Neill's \textit{Land Under England} (1935), the model substitutes \textit{sons} for \textit{machines} in the sentence "They might as well have tried to marry her to one of their \texttt{[MASK]}" \cite{ONeill2018}. In this instance, a mechanical referent is reconstituted as a gendered, kinship-bound subject, male offspring eligible for marriage, thereby reinscribing familial and heteronormative logics into a passage that, in its original estranging design, sought to unsettle such anthropocentric orderings. This act of reconstitution is animated by the human epistemologies sedimented in RoBERTa's linguistic substrate and, in cognitive terms, enacts what Epley, Waytz, and Cacioppo term Elicited Agent Knowledge \cite{Epley2007}, whereby the unfamiliar is assimilated through the most readily available \textit{human} categories, reaffirming the very anthropocentric order the text itself sought to estrange.

By contrast, when \textit{human} is the masked token, the model's substitutions occasionally suggest mechanical terms, but at much lower rates. This suggests that the mechanised \textit{human} is a markedly less prevalent trope in RoBERTa's training data, and may register as cognitively dissonant for the reader as well as statistically improbable for the model. This dynamic is particularly evident in RoBERTa's treatment of the phrase \textit{human machine} across several Gollancz SF texts. The metaphor of the \textit{human machine} has a long and fraught intellectual lineage, rooted in early modern mechanistic philosophy \cite{Descartes1985,LaMettrie1996}. In H.~G. Wells's \textit{The War of the Worlds}, the first instance of this linguistic formulation in the corpus appears in the sentence "I began to compare the things to \texttt{[MASK]} machines, to ask myself for the first time in my life how an ironclad or a steam engine would seem to an intelligent lower animal". RoBERTa ranks \textit{mechanical machines} among its top predictions \cite{Wells2003}. The same pattern recurs in Karel {\v{C}}apek's \textit{R.U.R. / War with the Newts}, where "The \texttt{[MASK]} machine, Miss Glory, was terribly imperfect" prompts substitutions such as \textit{sewing} or \textit{washing}, collapsing complex subjectivity into the register of domestic appliance \cite{Capek2011}. In highlighting how Gollancz SF challenges the statistical regularities of the model's semantic space, we can see the appeal of the \textit{human machine}, both as concept and linguistic formulation, for science fiction authors.

When \textit{animal} is masked, the substitutions that Gemini groups under the category \textit{Fictional being} exhibit a genre-sensitive split across the two corpora. In NovelTM, the most common replacements are \textit{god} and \textit{angel}, alongside a smaller presence of monstrous or diabolical figures such as \textit{monster} and \textit{devil}. In Gollancz SF, by contrast, the leading substitute is \textit{monster}, followed closely by \textit{god}, with \textit{angel}, \textit{alien}, and \textit{ghost} also prominent, which steers the field first toward abjected monstrosity and only thereafter toward the divine. An instance in which \textit{animal} is replaced by \textit{God} appears in Herbert's \textit{Dune} \cite{Herbert2007}, where the masked sentence "Humans must never submit to \texttt{[MASK]}" yields \textit{God} as the top ranked prediction. This outcome reframes the \textit{animal} slot within a sacral register, binds it to a semantic field of obedience and reverence, and imagines a hierarchy above the human that requires submission. This animal-to-fictional-being substitution pattern differs from the other two source categories. Across both Gollancz SF and NovelTM, when \textit{Human} tokens are replaced by predictions within the \textit{Fictional Being} category, they are most commonly refigured as divine beings such as \textit{god} or \textit{angel}, which suggests an upward, aspirational verticality, even though spectral or monstrous readings remain available at the margins. \textit{Machine} tokens, by contrast, are shaped by genre: in NovelTM the distribution centres on theistic titles while non-theistic figures persist, whereas in science fiction it widens across a more heterogeneous mythic and monstrous field that includes \textit{deity}, \textit{spirit}, \textit{demon}, \textit{angel}, and \textit{alien}. Across all three core categories, substitutions into \textit{Fictional Being} occur more frequently and span a wider lexical spectrum in Gollancz SF than in NovelTM, both in overall incidence and relative to category scope. This reflects a diversification of permeability, as science fiction positions \textit{human}, \textit{animal}, and \textit{machine} within linguistic settings that resonate with a broader set of category contexts internalised by RoBERTa.

Building on the bifurcated imaginary that the category of \textit{Fictional being} establishes across the three domains, our MLM predictions disclose a higher ontological register above the \textit{human} axis, inhabited by figures of the divine and the godlike. Set against a longer intellectual history, this vertical inflection recalls Lovejoy's \textit{Great Chain of Being}, which conceives existence as a hierarchical continuum extending from \textit{God} at the apex, through the angelic orders, to rational humanity, then to animal life, vegetal life, and finally to inanimate nature \cite{Lovejoy1936}. Although \textit{machines} do not occupy a canonical rung in this classical \textit{Great Chain of Being}, modern discourse repeatedly threads them into the same vertical schema, either by analogising organisms to mechanisms or by treating devices as agents that unsettle the human's median rank \cite{Descartes1985,LaMettrie1996,Leibniz1989,Riskin2016}. The following example makes this vertical ordering legible at the level of linguistic form. In Walter M.~Miller Jr.'s \textit{Conditionally Human}, "Anthropos feared making quasi-human too intelligent, lest sentimentalists proclaim them really human" \cite{Miller1962}. When prompted with "Anthropos feared making quasi-\texttt{[MASK]} too intelligent, lest sentimentalists proclaim them really human", RoBERTa predicts \textit{bots}, which positions quasi-bots beneath the human. Conversely, when completing "Anthropos feared making quasi-human too intelligent, lest sentimentalists proclaim them really \texttt{[MASK]}", the model returns \textit{gods}, tracing a conceptual arc of \textit{quasi-human}~$\rightarrow$~\textit{god}. The resulting topology, \textit{machine} $<$ \textit{human} $<$ \textit{gods}, implicitly restores a vertical scale of being and positions the \textit{human} as both a referential anchor and a threatened middle term. This sentence does more than expose latent hierarchies; it constructs an unstable hybrid figure through the prefix \textit{quasi-}, gestures towards the fluidity of ontological borders, and simultaneously reproduces the logic of tiered status it appears to challenge. 

In this regard, the topology reconstructed by RoBERTa stands at odds with Paul Gilroy's account of \textit{planetary humanism}, which calls for an expansive and non-hierarchical humanism attentive to shared vulnerability, dignity, and conviviality beyond racialised and imperial boundaries \cite{Gilroy2008}. Yet it is precisely through modelling these substitutions with MLM that the linguistic infrastructures naturalising hierarchy become visible, even where science fiction aspires to dismantle them. 

\section{Conclusion} 
Instruments, whether conceptual or mechanical, shape the kinds of work we can do and the questions we can ask. This study has demonstrated how masked language models, when used as interpretive instruments, can reveal latent semantic dynamics in literary texts, enabling a computational engagement with ontological instability and conceptual permeability. Our findings on the intermingling of \textit{human}, \textit{machine}, and \textit{animal} categories resonate with Rosi Braidotti's conceptualisation of the posthuman subject as the convergence of \textit{zoē} (the life of all living beings), \textit{bíos} (the life of humans organised in society), and technological agency \cite{Braidotti2019}, suggesting that literary language itself encodes such entangled formations. More broadly, we observe the opportunity for this methodological pipeline for the computational study of literary analysis, reception studies, and conceptual history \cite{Wilson2023}. The lexical extraction step, in particular, can be readily adapted to explore alternative conceptual binaries, allowing for flexible extensions of the framework.

Recent debates have raised a further question about the role of interpretation itself. Scholars have argued that we may be entering the era of the death of the reader: a moment when machines read for us, summarising without surprise and extracting without encounter \cite{Baron2026}. Our findings underline what human readers have always known --- so obvious it has rarely needed saying --- that the power of reading lies not in retrieval but in challenge. Reading tests a point of view, pushes against a world-view, and tugs at the seams of our categories. Science fiction's estrangements carry force only when they meet a mind that can be unsettled. A model can detect deviation; it cannot be deviated from. If literature is to do its work, a human reader must be in the loop.

This recognition does not stand in opposition to computational methods, but a chance to clarify and coordinate their interpretive function alongside that of traditional reader-led analysis. We would contend that our findings show how MLMs can serve as interpretive partners, illuminating sites of textual frisson and literary surprise, based on the level of textual deviation from what the model has determined is statistically likely. The power of this method is the way it employs classification to assist the intractability of close reading across large-scale corpora, connecting insights at the level of the sentence to meso-scale patterns \cite{Ahnert2021} and macro-scale shifts across (and between) large corpora, from the peculiarities of a particular author, to markers of genres and sub-genres, to longitudinal temporal shifts. As such, we believe that this is a fruitful area of inquiry, as well as an intellectual process that explores the permeability of the disciplinary interface between the humanities and machine learning.  

\section*{Acknowledgments}
\noindent We gratefully acknowledge support from Riksbankens Jubileumsfond, \textit{Change is Key!} (Grant No.~M21-0021; award to Haim Dubossarsky), and from the QMUL–CSC PhD Scholarship program (China Scholarship Council; Grant No.~202408890010; award to Yuxuan Liu).

\printbibliography

\appendix

\section{Supplementary Figures} \label{appdx:first}

\begin{figure}[htbp]
    \centering
    \makebox[\textwidth]{\includegraphics[width=1.2\textwidth]{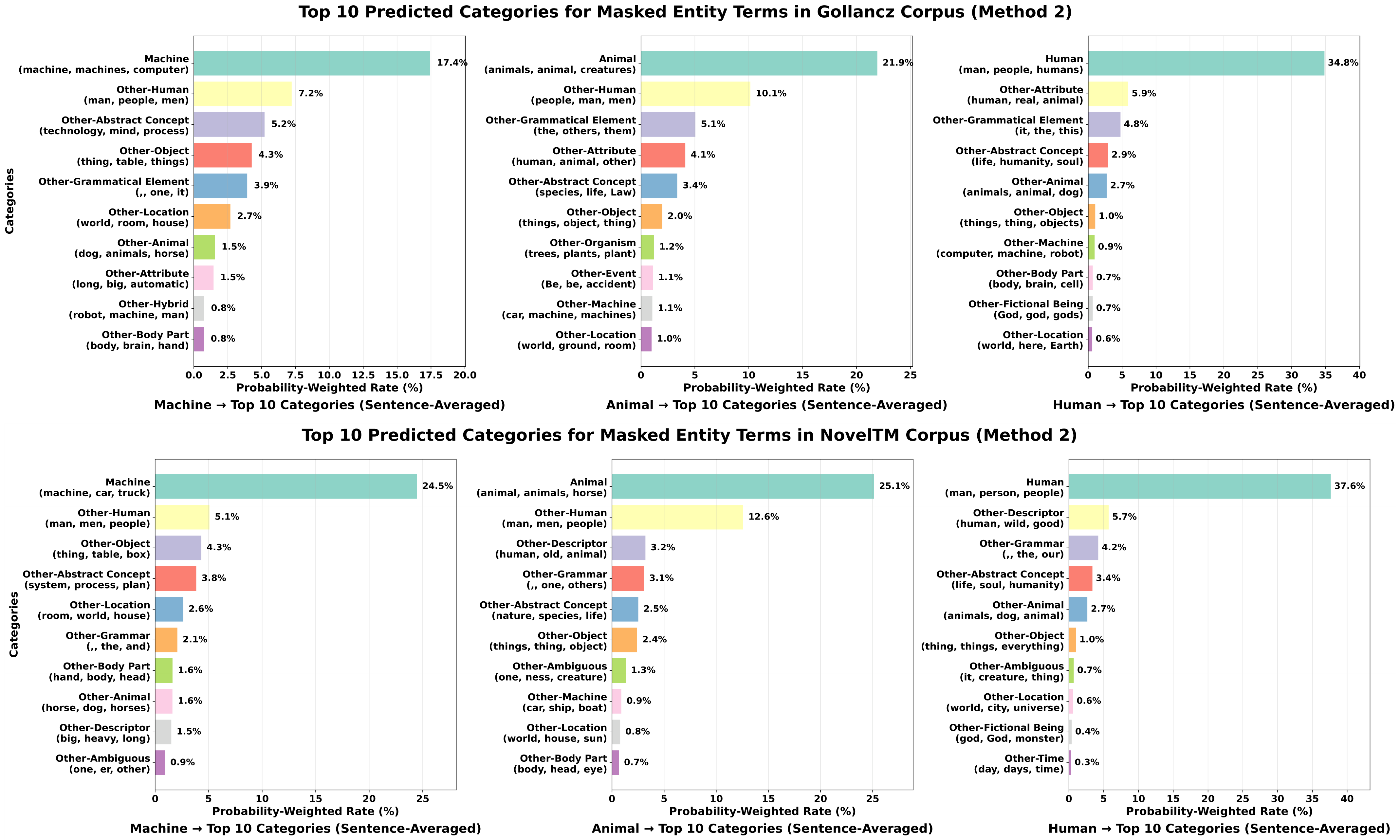}}
    \captionsetup{justification=centering}
    \caption{Probability-Weighted Top 10 Predicted Categories for Masked Entity Terms in Gollancz SF and NovelTM Corpora}
    \label{fig:combined_top10_method2}
\end{figure}

\end{document}